\documentclass[letterpaper,10pt,conference]{IEEEtran}  

\pdfoutput=1

\usepackage{xparse} 
\usepackage{amsmath} 
\usepackage{amssymb}
\usepackage{amsfonts} 
\usepackage{bbm}
\usepackage{color}
\usepackage{verbatim}
\usepackage{multirow}
\usepackage[flushleft]{threeparttable}
\usepackage{graphicx}
\usepackage[font=small]{caption}
\usepackage{balance}

\usepackage{cleveref}
\usepackage[font=small,position=b]{subcaption}
\usepackage{color}
\definecolor{mygray}{gray}{0.4}
\usepackage{float}
\usepackage{caption}
\usepackage{cuted}

\voffset=\baselineskip

\IEEEoverridecommandlockouts
\IEEEpubid{\makebox[\columnwidth]
	{\hspace{0.4em}978-1-5090-6299-7/17/\$31.00~\copyright{\hspace{0.15em}2017} IEEE.\hfill}%
	\hspace{\columnsep}\makebox[\columnwidth]{ }}

\usepackage{fancyhdr}
\fancypagestyle{firstpage}{
	\fancyhf{}
	\fancyhead[C]{\textcolor{mygray}{2017 International Conference on Indoor Positioning and Indoor Navigation (IPIN), 18--21 September 2017, Sapporo, Japan}}
	\makeatletter
	\fancyfoot[L]{%
		\normalfont\footnotesize%
		\raisebox{\footskip}[1.5ex]{%
			~~~~}}
	\makeatother %
}

\usepackage{url}
\usepackage{framed}

\usepackage{booktabs}
\usepackage{arydshln}

\newcommand\T{\rule{0pt}{2.6ex}}        
\newcommand\B{\rule[-1.2ex]{0pt}{0pt}}  

\NewDocumentCommand\Matrix{m}{ \boldsymbol{\mathbf{#1}} }

\newcommand{\mbf}[1]{\mathbf{#1}}

\DeclareMathAlphabet{\mbfh}{OML}{cmm}{b}{it}

\newcommand{\cframe}[1]{\ensuremath \underrightarrow{\mathcal{F}}_{#1}}

\newcommand{\norm}[1]{\left\Vert#1\right\Vert}

\newcommand{\bbm}{\begin{bmatrix}}
\newcommand{\ebm}{\end{bmatrix}}

\newcommand{\SO}[1]{SO(#1)}

\newcommand{\SE}[1]{SE(#1)}

\title{Improving Foot-Mounted Inertial Navigation\\ Through Real-Time Motion Classification}
\author{Brandon Wagstaff, Valentin Peretroukhin, and Jonathan Kelly\thanks{All authors are with the Space \& Terrestrial Autonomous Robotic Systems (STARS) Laboratory at the University of Toronto Institute for Aerospace Studies (UTIAS), Canada {\tt \{brandon.wagstaff, valentin.peretroukhin\}@robotics.utias.utoronto.ca, jkelly@utias.utoronto.ca}}}

\begin{document}
\maketitle 
\thispagestyle{firstpage}

\begin{abstract}
We present a method to improve the accuracy of a foot-mounted, zero-velocity-aided inertial navigation system (INS) by varying estimator parameters based on a real-time classification of motion type.  We train a support vector machine (SVM) classifier using inertial data recorded by a single foot-mounted sensor to differentiate between six motion types (walking, jogging, running, sprinting, crouch-walking, and ladder-climbing) and report mean test classification accuracy of over 90\% on a dataset with five different subjects.

From these motion types, we select two of the most common (walking and running), and describe a method to compute optimal zero-velocity detection parameters tailored to both a specific user and motion type by maximizing the detector F-score. By combining the motion classifier with a set of optimal detection parameters, we show how we can reduce INS position error during mixed walking and running motion. We evaluate our adaptive system on a total of 5.9 km of indoor pedestrian navigation performed by five different subjects moving along a 130 m path with surveyed ground truth markers. 
 \end{abstract}

\section{Introduction}

Localization within indoor environments can often be challenging because building materials can significantly attenuate or reflect GNSS-based navigation signals. For indoor pedestrian tracking, one potential alternative is body-mounted inertial navigation, a dead-reckoning approach that integrates inertial rates to estimate the position, orientation and velocity of a moving person within a predefined coordinate frame. 
For low-cost inertial measurement units (IMUs) based on  microelectromechanical systems (MEMS), relying on direct integration for any appreciable interval is impractical, as position error grows cubicly with time \cite{ZVDetect}. To bound this error growth, a zero-velocity-aided inertial navigation system (INS) uses periodic zero-velocity updates (ZUPTs). ZUPTs occur during \textit{midstance}, a part of the human gait in which the foot is stationary relative to the ground.  
By mounting an IMU to the foot of the user, a zero-velocity event can be detected through a general likelihood ratio test (LRT) on the inertial data \cite{zupteval,ZVDetect}, and then incorporated into an extended Kalman filter (EKF) as a pseudo-measurement.

\begin{figure}[t]
	\small
	\centering
	\begin{subfigure}[]{0.48\textwidth}
	\includegraphics[width=\textwidth]{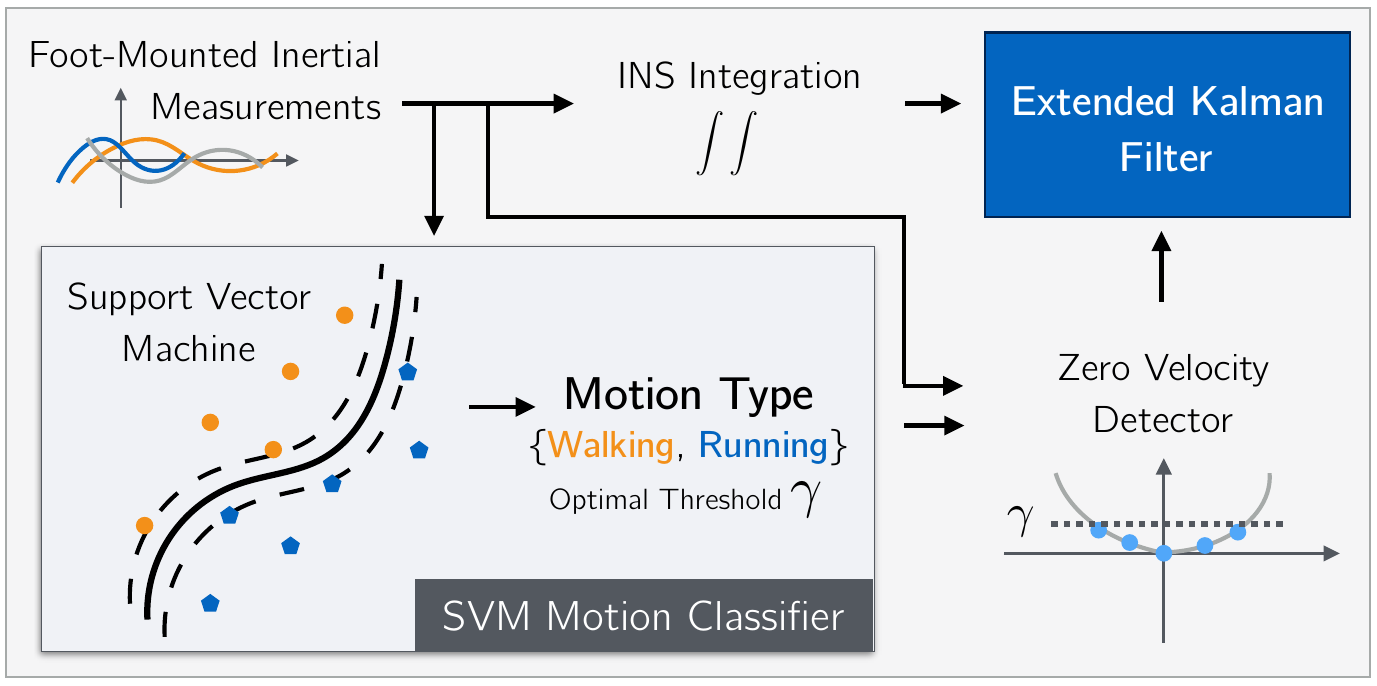}
	\caption{Our proposed system.}
	\vspace{0.5em}
	\label{fig:system}
	\end{subfigure}
	~
	\begin{subfigure}[]{0.48\textwidth}
		\includegraphics[width=\textwidth]{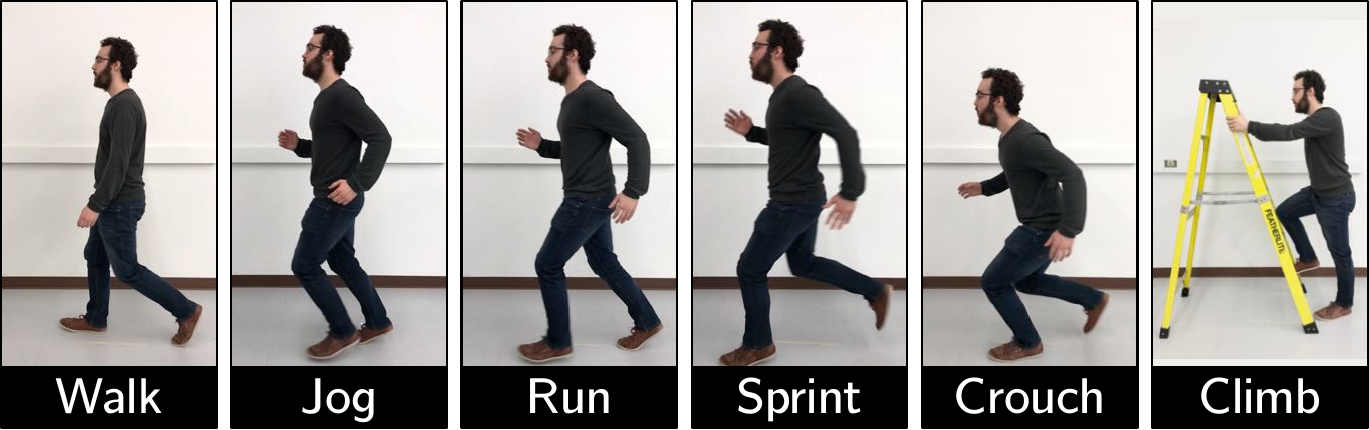}
		\caption{Six different motion types. } 
		\label{fig:motion_types}
	\end{subfigure}
	\caption{(Top) Our system consists of an SVM that classifies motion based on inertial data. We use the motion class to select optimal parameters for a zero-velocity detector. (Bottom) Six different motion types that we use to train and test our SVM-based motion classifier.} \vspace{-0.5em}
	\end{figure}

%

In this work we revisit a known drawback of ZUPT-based filters, namely that the optimal zero-velocity detection parameters are dependent on motion type \cite{ZVDetect}. First, for a fixed motion class,  we outline a method to compute an optimal zero-velocity detection threshold that balances both the precision and recall of the detector. Second, we describe a motion classifier based on a support vector machine (SVM) to accurately classify motion in real-time from inertial data. Although we focus on walking and running motion in this paper, we show that our classifier is powerful enough to classify six different motion classes (\Cref{fig:motion_types}) with accuracies exceeding 90\%. Finally, we combine our classifier with a set of optimal zero-velocity thresholds to create a more robust navigation system that can track dynamic, high-speed motions while maintaining high accuracy during walking.  \Cref{fig:system} illustrates our proposed system. In short, the main contributions of this work are:
\begin{enumerate}
\item a procedure to determine optimal zero-velocity detection parameters for a range of motions,
\item a real-time capable classification routine that can reliably distinguish between different motions from a single foot-mounted IMU, and
\item an evaluation of an adaptive, classification-based zero-velocity-aided INS using a substantial 5.9 km indoor navigation dataset with surveyed ground truth markers and involving five different subjects.
\end{enumerate}
\newpage
\section{Related Work}
Inertial sensors have a rich history in assisting the navigation of ships, airplanes, and spacecraft throughout the twentieth century. While early gyroscopes and accelerometers were large, the availability of lightweight, low-cost, MEMS-based sensors has enabled the production of commercial IMUs for pedestrian navigation.  Foxlin et al.\ \cite{Foxlin:2005} first introduced a zero-velocity-aided, foot-mounted INS that used ZUPT detection to bound position error.\footnote{Zero-velocity measurements can correct for position, velocity, accelerometer biases, pitch, roll, and the pitch and roll gyro biases. The yaw and yaw gyro bias remain unobservable during the zero-velocity update.}  Midstance detection was studied in detail by Skog et al.\ \cite{zupteval,ZVDetect} who derived four zero-velocity detectors in the general LRT framework. 

Nilsson et al.\ \cite{INS_characterization:2010,zupteval} demonstrated that the parameters for each LRT (e.g., window size, noise variance, and hypothesis threshold)  can be selected to minimize error along a trajectory for a specific user.  In further work \cite{Nilsson:2012}, Nilsson showed that the higher hypothesis thresholds required for midstance detection during faster motion lead to increased error during walking.  Rantakokko et al.\ \cite{Rantakokko:2012} reported that both movement and surface type `strongly influenced the performance' of a zero-velocity-aided INS, and recommended `more robust stand-still detection algorithms' for movements such as sprinting, jogging, sidestepping, ascending and descending stairs, and crawling. Additional work by Nilsson et al.\ \cite{Nilsson2:2014} also reported large error accumulation in INS estimates for crawling users (firefighters).  
 
\subsection{Adaptive Thresholding}

To create more robust zero-velocity detectors that work reliably across a range of motion types, several adaptive approaches have been presented in the literature. These techniques vary zero-velocity detection thresholds based on some characterization of the foot state.  For instance, Walder and Bernoulli \cite{context-adaptive} developed a context-adaptive algorithm that aimed to detect midstance during common motion patterns such as walking, running, and crawling by using a velocity-dependent thresholding algorithm.  In a similar manner, Ren et al.\ \cite{Sensors2016} introduced a velocity-based detector that used a state machine to transition between motion and zero-velocity states, with the transition probability governed by estimated velocity. They argued that their detector performed better than gyroscope-based detection during non-walking motions because angular velocity rates are non-zero at midstance when the user is moving quickly. Li and Wang \cite{Li:2012} presented a zero-velocity detection algorithm that worked during both walking and running motion by computing two individual zero-velocity detections, each tuned specifically for its motion type. Finally, Tian et al.\ \cite{Sensors2016_2} extracted the user's gait frequency from the IMU signal and used it to adaptively update the zero-velocity threshold. 

We note that the majority of these methods adapt parameters of a single detector based on foot velocity or gait frequency. While this paradigm may be effective for smooth, continuous motions such as walking and running,  it can fail to correctly account for other motion types such as stair-climbing, or crawling (where angular rates, which have been shown to be critical to detect midstance during walking \cite{ZVDetect}, may not play a significant role).  Indeed, selecting a single zero-velocity detector that can robustly identify midstance across a range of motions is a difficult task. 

\subsection{Motion Classification}

 To overcome the challenge of achieving robust midstance detection across a range of motions, our work focuses on classifying motion to facilitate midstance detection. Given a motion type, we can select a detector that is optimized for that particular motion and obviate the need to burden a single detector with multiple motion types. Several machine learning techniques have been used for motion classification in the literature. Mannini and Sabatini \cite{Mannini2010} compared a variety of classification methods (Naive Bayes, logistic regression, nearest neighbour, and an SVM) to identify motion from several body-mounted sensors. 
Lau et al.\ \cite{svmclass} trained an SVM to predict motion types using data acquired from gyroscopes and accelerometers placed on the shank and foot. Their approach used inertial data collected from test subjects performing five walking motions: level-ground walk, upslope, downslope, stair ascent and stair descent. They reported a classification accuracy of 84.71\% for a five-motion classifier, and 100\% for two and three-motion classification.  

Using only foot-mounted inertial sensors, Park et al.\ \cite{Park2016} implemented a zero-velocity detection system that used two SVMs: one to classify a user's motion, and another to identify midstance events.  They reported midstance detection accuracies greater than $99\%$, but did not report localization improvements. In our work, we adopt a similar, SVM-based approach to motion classification. However, our system uses classification to modify parameters of an existing LRT-based detector, instead of implementing an entirely new learned mechanism.

\section{System Overview}
\label{sec:overview}

In the sections that follow, we present our zero-velocity-aided INS, which incorporates a novel motion classifier for improved midstance detection.

\subsection{Zero-Velocity Aided INS with an Extended Kalman Filter}
For our baseline INS, we use an EKF to track the state of a foot-mounted IMU. The filter's nominal state\footnote{We do not incorporate any sensor biases into the state, as Nilsson et al. \cite{Nilsson:2012} note that doing so will not improve the accuracy of the system since the modelling error associated with zero-velocity measurements significantly outweighs error due to sensor bias.} consists of the IMU's position ($\mbf p_k$), velocity ($\mbf v_k$), and orientation in quaternion form ($\mbf q_k$),
\begin{align}
\label{eq:state}
\mathbf{x}_k &= \bbm \mbf p_k \\ \mbf v_k  \\ \mbf q_k \ebm.
\end{align}
To propagate the nominal state, the EKF uses a non-linear motion model $f(\cdot)$ given IMU inputs $\{\mathbf{a}_k^b, \boldsymbol{\omega}_k^b\}$\footnote{$\mathbf{a}_k^b$  and $\boldsymbol{\omega}_k^b$ are the three-axis acceleration and angular velocity expressed in the IMU (body) frame.},
\begin{align}
\label{eq:naveq}
\mbf{x}_k &= f(\mbf{x}_{k-1},\mbf{a}_k^b, \boldsymbol{\omega}_k^b) \nonumber \\ 
&= 
\bbm
\mbf{p}_{k-1} + \mbf{v}_{k-1}\Delta t \\ 
\mbf{v}_{k-1} + \left( \mbf{R}(\mbf{q}_{k-1})\mbf{a}_k^b - \mbf{g} \right)\Delta t  \\
\mbf{\Omega} (\boldsymbol{\omega}_k^b \Delta t)\mbf{q}_{k-1} \\
\ebm,
\end{align}

\noindent where $k$ is a time index, $\Delta t$ is the sampling period, $\mathbf{R}(\cdot) $ is a function that maps quaternions to \SO{3} matrices, $\mathbf{g}$ is the gravity vector, and $\mbf{\Omega}(\boldsymbol{\phi})$ is a 4x4 matrix that updates the quaternion state based on an incremental rotation $\boldsymbol{\phi}$, which we compute through a backward, zeroth order integration of angular-rates (i.e., $\boldsymbol{\phi}=\boldsymbol{\omega}_k^b \Delta t$).  Alongside the nominal state, the filter maintains a minimal error state and uses it to apply corrections when a midstance event is detected. During midstance, the filter incorporates a direct observation of velocity and fuses this pseudo-measurement with the motion model according to the standard EKF framework. For a more detailed explanation of an EKF and a zero-velocity-aided INS we refer the reader to \cite{Foxlin:2005,Sola:2016}.

For midstance detection, we choose to use stance hypothesis optimal detection (SHOE) for its robustness to changes in gait speed and high positional accuracy \cite{zupteval}. Intuitively, the SHOE detector thresholds the sum of the energy of angular rates and of linear accelerations with gravity `removed', over a window of $W$ measurements. To obviate the need for a global orientation estimate, gravity is `removed' by subtracting a vector of magnitude $g$ in the direction of average acceleration \cite{ZVDetect}. Concretely, SHOE tracks when the statistic $T_n(\mathbf{a}^b, \boldsymbol{\omega}^b)$, falls below $\gamma$:   

\vspace{-0.3cm}
\begin{equation}
	\label{eq:shoe_detector}
	\begin{split}
		T_n(\mathbf{a}^b, \boldsymbol{\omega}^b) &= \\
 \frac{1}{W} \sum_{k=n}^{n+W-1}  & \left( \frac{1}{\sigma_a^2}\norm{ \mathbf{a}_k^b - g\frac{\bar{\mathbf{a}}_n^b}{\norm{\bar{\mathbf{a}}_n^b}}}^2 + \frac{1}{\sigma_\omega^2}\norm{\boldsymbol{\omega}_k^b}^2 \right) < \gamma,
	\end{split}
\end{equation}
\vspace{-0.1cm}

\noindent where $W$ is the window size (the number of sensor readings the detector observes), $\sigma_a^2, \sigma_\omega^2$ are the variances of the specific force and angular rate measurements, $\mbf {\bar{a}}_n^b$ denotes the sample mean over $W$ samples, and $g$ is the magnitude of the local gravitational acceleration.

\subsection{Adaptive Zero-Velocity Detector with Motion Classification}
\label{sec:realtimeINS}

\begin{figure}[t]
	
	\includegraphics[width=0.48\textwidth]{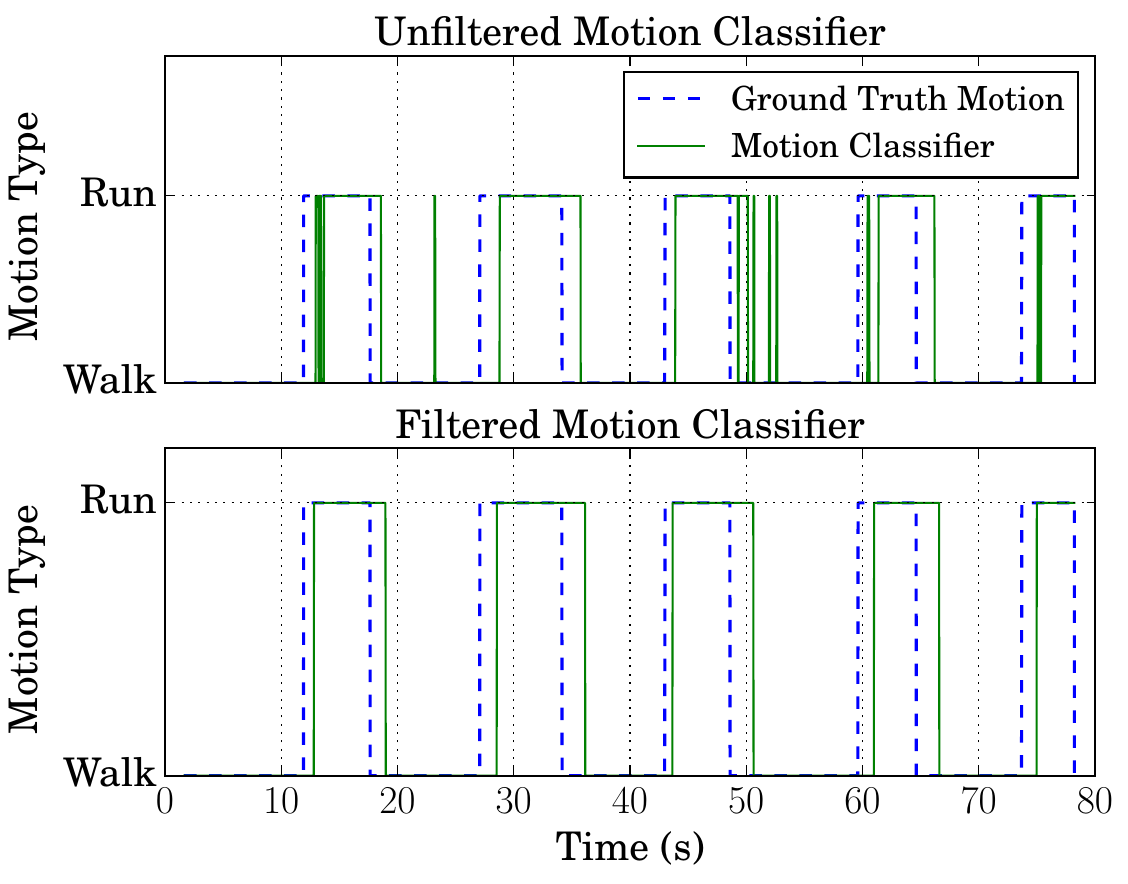}
	\caption{Unfiltered (top) and filtered (bottom) classification results with associated ground truth.}
	\label{fig:motionclass}
\end{figure}

In this paper, we introduce an adaptive zero-velocity detector that uses an SVM classifier to determine the user's motion type. SVMs are linear classifiers that use a hyperplane to maximally separate groups of differently labelled data.  Contrary to other linear classifiers, SVMs can typically classify datasets that are not linearly separable by first transforming them to a higher dimensional feature space where they become linearly separable.  Our classifier is trained with inertial data that is labelled with the user's motion type (see \Cref{sec:motionclassification}).  \Cref{fig:motionclass} illustrates the motion classifier's ability to distinguish walking from running.  We post-process the classifier output using a mean filter to remove abrupt motion transitions.

Given a motion class, our zero-velocity detector selects an optimal parameter set for the current motion.  The entire adaptive system operates in real-time using the Robot Operating System (ROS), displaying the estimated trajectory in RViz, a visualization tool \cite{rospaper}.  

\section{Experiments}

We outline and describe the validation of our proposed system in three sections. First, in \Cref{sec:opt_thresh}, we present a method to select optimal parameters of the zero-velocity detector, tailored to a specific motion type and user.  In \Cref{sec:motionclassification} we present our SVM-based motion classification in detail, and evaluate its accuracy with six motion classes.  Finally, in \Cref{sec:adaptive_ins_eval} we combine the classifier with a set of optimal parameters and evaluate the adaptive INS using an indoor dataset.

\label{sec:experiments}
\subsection{Determining Optimal Thresholds for Midstance Detection}
\label{sec:opt_thresh}
To select optimal parameters for a zero-velocity detector, we generate zero-velocity ground truth by tracking the motion of a subject's foot using a Vicon infrared motion tracking system.  While these parameters can be tuned indirectly by minimizing position error over a trajectory, we optimize zero-velocity events directly to avoid potential dependencies between position error and the geometry of a particular trajectory \cite{INS_characterization:2010}.  In this work, we use the SHOE detector, fixing the parameters $W$ and $\frac{\sigma_a}{\sigma_\omega}$, while focusing our attention on $\gamma$.\footnote{We note that the threshold affects motion-specific detection more than the other tuning parameters do.  While tuning other parameters may improve detection in non-walking motions, we leave this as future work. Our fixed parameters were: $W = 5$, $\sigma_a=0.01$, and $\sigma_\omega =0.00174$.}

\begin{figure}[t]
	\centering
	\includegraphics[width=0.48\textwidth]{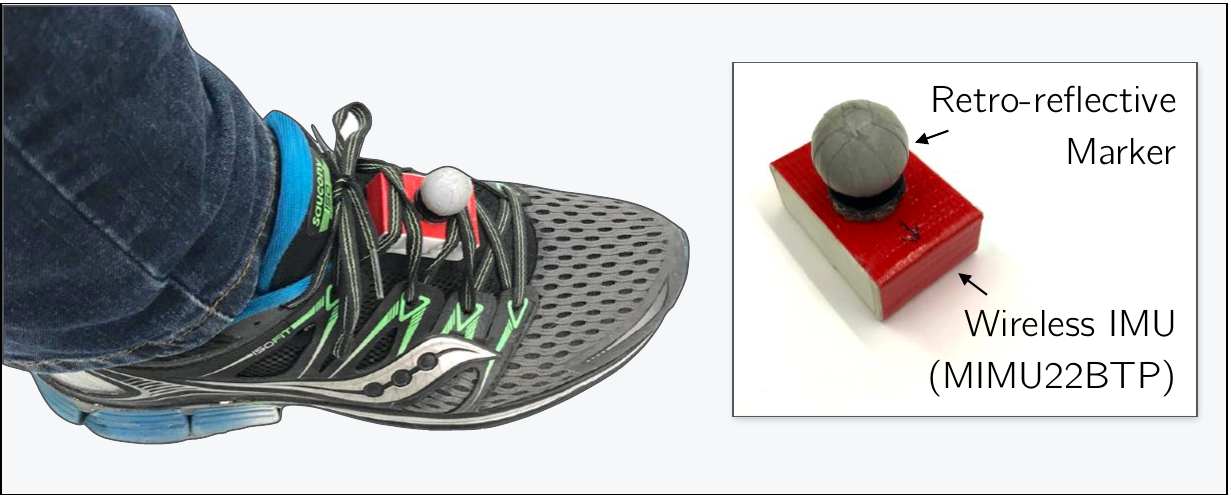}
	\caption{Our apparatus: an Inertial Elements MIMU22BTP(X) \cite{Gupta:2015}  wireless 4-IMU array mounted to the foot of a test subject. Note that we use both the MIMU22BTP and its X variant (which has a larger lithium-ion battery but identical sensing hardware). }
	\label{fig:foot_imu}
\end{figure}

\subsubsection{Data Collection} 
\label{sec:param_opt}

To record training data, we mounted a wireless IMU to the foot of five different subjects. In all experiments presented in this paper, we used the Inertial Elements Osmium MIMU22BTP(X) (a wireless, Bluetooth-enabled MEMS-based 4-IMU sensor array \cite{Gupta:2015}). We relied on the internal processing of the sensor to fuse the four sets of inertial measurements into a single 6-axis reading, operating at 125 Hz.  To observe the ground truth position of the IMU, we attached a Vicon marker  to the sensor itself (see \Cref{fig:foot_imu}) and recorded both the inertial data and Vicon motion tracking using ROS.  All subjects wore their preferred pair of running shoes, and the IMU was mounted approximately in the centre of the foot using the shoe's laces.  Each user walked or ran for 10 laps within the Vicon tracking volume. A sample trajectory is shown in \Cref{Vicontraj}.  In practice, we found that we only needed 1--2 laps of data (approximately 20 seconds) to extract optimal parameters.

By numerically differentiating the Vicon position data, we computed foot speeds and applied a threshold to generate ground truth zero velocity events (see \Cref{zvgt}). We used thresholds of 0.1 and 0.25 m/s for walking and running, respectively. The thresholds were empirically selected to ensure the ground truth captured every midstance event without exaggerating their length.

\subsubsection{F-Score Optimization} 

\begin{figure*}[t]
	\begin{subfigure}[b]{0.49\textwidth}
		\includegraphics[width=\textwidth]{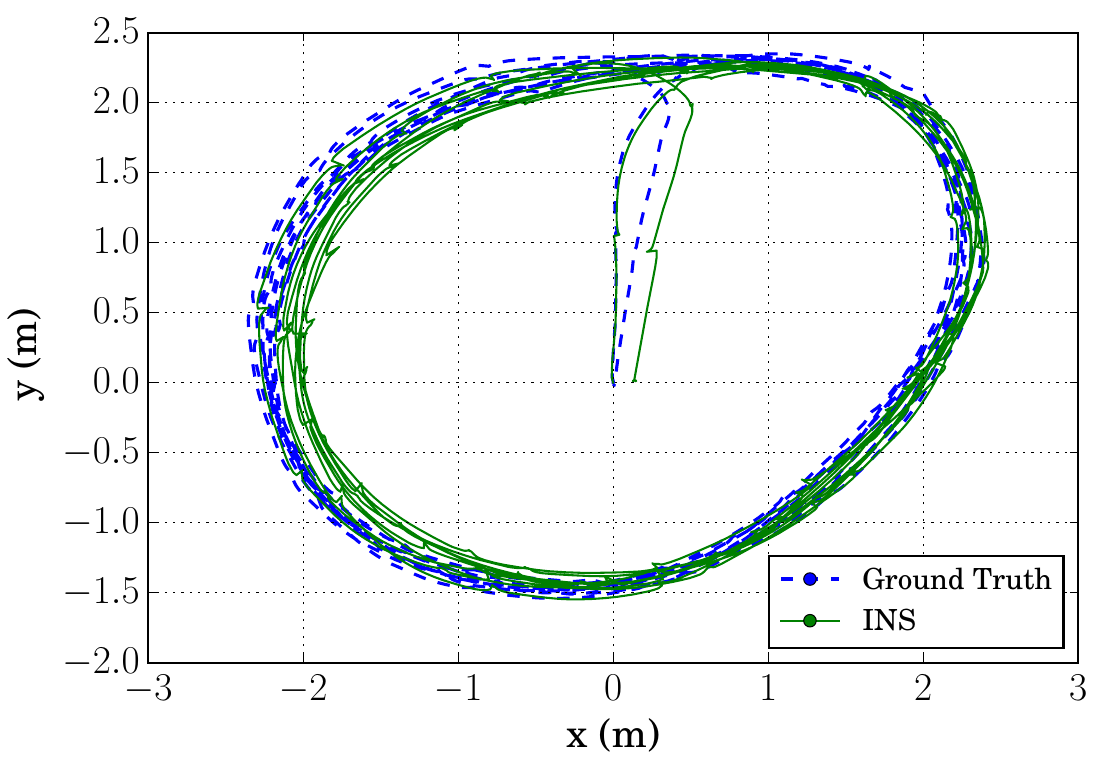}
		\caption{Top-down projection of a walking trajectory within the Vicon motion capture volume.  Ground truth estimates are shown along with the output of a zero-velocity-aided INS using an optimized, but fixed, $\gamma_{walk}$.  }
		\label{Vicontraj}
	\end{subfigure}
	~
	\begin{subfigure}[b]{0.49\textwidth}
		\includegraphics[width=\textwidth]{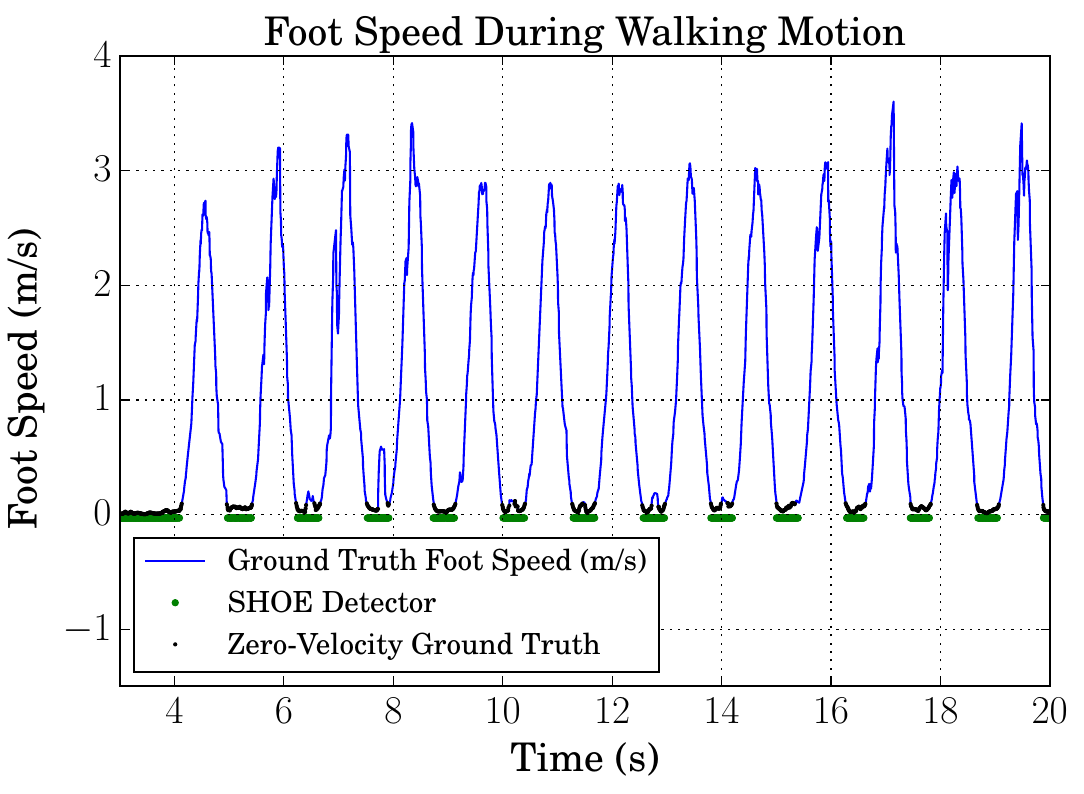}
		\caption{Foot speed computed by numerically differentiating Vicon position data while a user (subject 5) is walking.  We define midstance when foot speed is below 0.1 m/s for walking and 0.25 m/s for running.}
		\label{zvgt}
	\end{subfigure} \\
	\begin{subfigure}[t]{0.49\textwidth}
		\includegraphics[width=\textwidth]{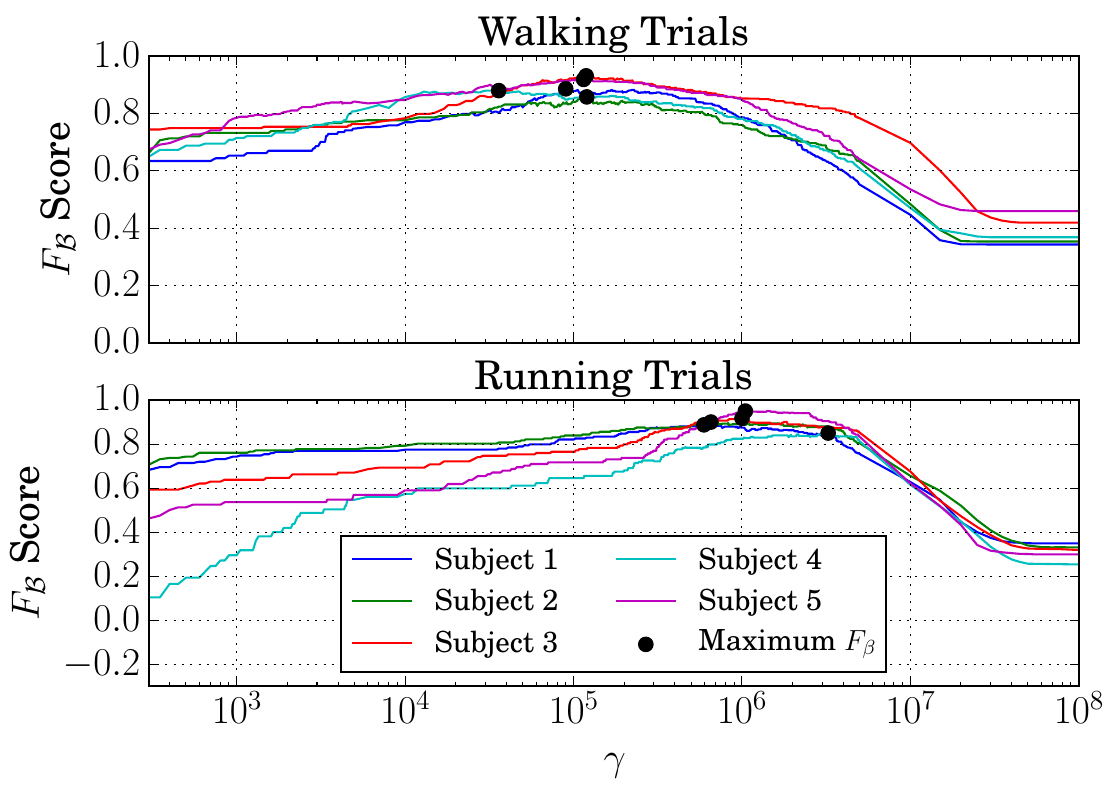}
		\caption{$F_\beta$ scores for a range of $\gamma$, with maximum points for all subjects' walking and running trajectories.}
		\label{fscore}
	\end{subfigure}
	~
	\begin{subfigure}[t]{0.49\textwidth}
		\includegraphics[width=\textwidth]{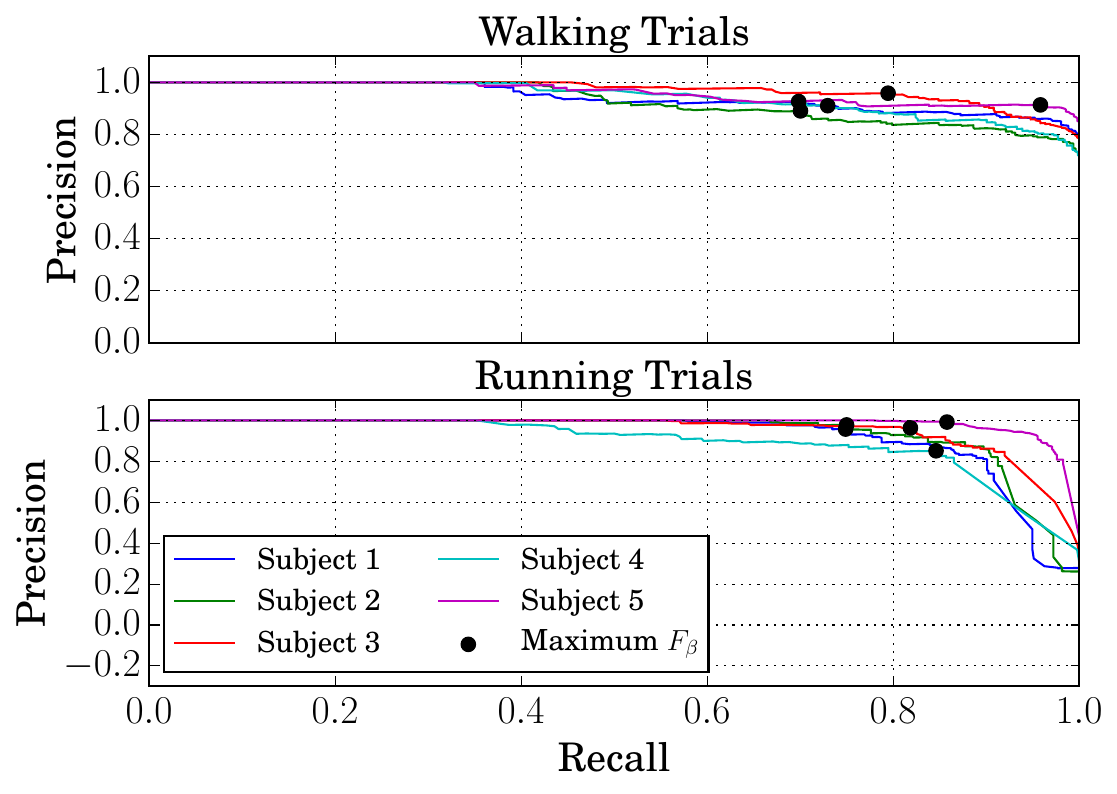}
		\caption{Precision-recall curves for all subjects' walking and running trials.  The optimal $\gamma$ (corresponding to the maximum $F_{\beta}$ score) locations are shown.}
		\label{precrec}
	\end{subfigure}
	\caption{Optimizing the zero-velocity detector threshold from ground truth position data.}
	\vspace*{-2mm}
\end{figure*}

We compared SHOE detector output with zero-velocity ground truth while varying $\gamma$ through the range $[10^2, 10^8]$. At each value of $\gamma$, we computed the precision ($P$) and recall ($R$) to form a precision-recall curve (\Cref{precrec}).  To select an optimal operating point, we maximized the $F_\beta$ score (\Cref{fscore}):

\vspace{-0.2cm}
\begin{equation}
F_\beta = \left(1+\beta^2\right)\frac{P R}{\beta^2 P + R}.
\end{equation}
\vspace{-0.3cm}

 In this F-measure, the $\beta$ parameter controls the importance of precision relative to recall. For $\beta < 1$, precision is favoured over recall, and decreasing $\beta$ generally moves the operating point to the left on the precision-recall curve. Empirically, we found that precision was slightly more important for walking compared to running, though both regimes required $\beta < 1$. In this paper, we use $\beta^2$ values of 0.16 and 0.4 for walking and running respectively, and leave a further investigation into potential detector trade-offs for future work. 

\subsubsection{Results} 

Table \ref{optimal_gammas} presents the $\gamma$ values that correspond to the maximum $F_\beta$ score for each user.  Note that in all cases, the walking threshold is significantly smaller than the running threshold, as expected.  

\begin{table}[]
	\centering
	\caption{Optimal walking and running zero-velocity thresholds ($\gamma$), found by maximizing the $F_\beta$ score using the Vicon zero-velocity ground truth.}
	\label{optimal_gammas}
	\begin{tabular}{llccccc | l}
		Subject                                           &      & 1    & 2   & 3   & 4    & 5   & Mean          \T \\ \midrule
		Optimal $\gamma$ & Walk & 0.90  & 1.20 & 1.19 & 0.36 & 1.15 & \textbf{0.96} \\
		($\times 10^5$)  & Run  & 5.96 & 6.55 & 10.02 & 32.55 & 10.49 & \textbf{13.11} \\ \bottomrule \B
	\end{tabular}
	\vspace*{-5mm}
\end{table}

\subsection{Motion Classification}
\label{sec:motionclassification}
To select which set of optimal parameters to use for zero-velocity detection, we must have some notion of motion type. In our work, we rely on a real-time classification of motion using an SVM classifier. Here, we discuss the training and evaluation of this classifier in more detail.

\subsubsection{Data Collection} 

We collected foot-mounted inertial data from five people, who each recorded six separate motion trials that consisted of either walking, jogging, running, sprinting, crouch-walking, or ladder-climbing (see \Cref{fig:motion_types}).  For each trial, the users moved along a circular trajectory (with a radius of approximately 3 m) for 10 laps.\footnote{For ladder climbing, each subject ascended and descended a step ladder ten times.} 

\subsubsection{Training} 

To train the classifier, we pre-processed the motion trials in several steps. First, we removed 1000 data points (approximately 8 seconds of inertial data at 125 Hz) from the beginning and end of each trial to ensure that each trial consisted of pure motion data.  Next, we normalized the three gyroscope and accelerometer channels of the IMU to ensure that they had similar magnitudes.  We separated the normalized IMU samples into training and test sets, using the first half of each motion trial for training, and the second half for evaluation of a test set.

Given the IMU specific force ($\mbf a_k$) and angular velocity measurements ($\boldsymbol{\omega}_k$) at timestep $k$,

\vspace{-0.3cm}
\begin{equation}
\mbf a_k = [a_k^x, a_k^y, a_k^z],
\end{equation}
\begin{equation}
\boldsymbol{\omega}_k = [\omega_k^x, \omega_k^y, \omega_k^z],
\end{equation}
we selected $K = 125$ (i.e. 1 second of data) adjacent IMU timesteps to form a training sample ($\mbf d_i$):
\begin{equation}
\mbf d_i = [\mbf a_k, \boldsymbol{\omega}_k, \dots, \mbf a_{k+K-1}, \boldsymbol{\omega}_{k+K-1}].
\end{equation}
Given a sample, $\mbf d_i$, we designed our SVM classifier ($g(\mbf d_i$)) to output a predicted motion type ($y_i$):
\begin{equation}
g(\mbf d_i) = y_i \in \{0,1, \dots, 5\},
\end{equation}
\vspace{-0.3cm}

\noindent where the integers represent walking, jogging, running, sprinting, crouch-walking, and ladder-climbing respectively. For each motion type, we combined 1000 samples from each of the five users. Our entire dataset consisted of 30,000 training and 30,000 test samples in six motion classes (5000 samples per class). Using the Radial Basis Function (RBF) kernel, we trained the SVM with the six motion classes from all five subjects' training sets.  

\subsubsection{Results} 
\begin{figure}[t]
    \centering
      \includegraphics[width=0.48\textwidth]{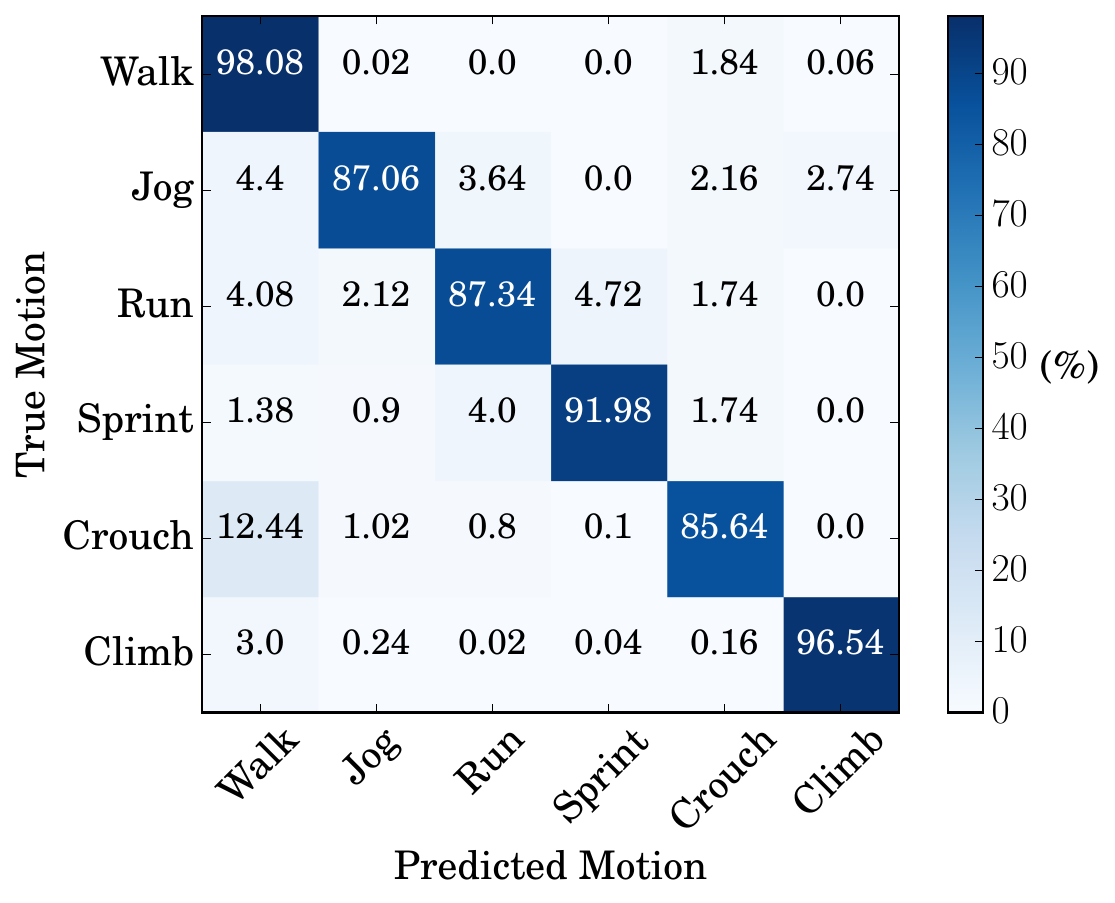}
      \caption{Confusion matrix depicting SVM Motion classification output.}
    \label{fig:svm_confusion_matrix}
\end{figure}

The confusion matrix in \Cref{fig:svm_confusion_matrix} shows the test set's classification accuracies for each motion type (averaged over the five subjects). The average motion classification rate was $91.1\%$.  When simplifying to a binary classifier (walk vs. sprint) we achieved accuracies above $99.9\%$.

\subsection{Adaptive, Classification-Based INS}
\label{sec:adaptive_ins_eval}

Combining our classifier with a set of optimal zero-velocity parameters, we evaluated the position accuracy of our adaptive INS to that of a fixed-threshold approach.   For this, we limited our motion classification to the binary case of walking and running, and collected an extensive indoor dataset.

\subsubsection{Ground Truth Acquisition} 

We created a ground truth trajectory through our hallway environment at the University of Toronto Institute for Aerospace Studies (UTIAS) by surveying six floor markers spaced at approximately 15~m intervals using a Leica Nova MS50 MultiStation (see \Cref{fig:hallway}).  Each marker consisted of an AprilTag \cite{apriltag} of 28 cm side length affixed to the floor.  Although AprilTags were originally designed as a visual fiducial system, we used only their outline to define the orthogonal coordinate axes of each marker frame, with the bottom left corner of the AprilTag defining the marker itself. We note that the mapped AprilTags can also be used, in principle, to provide high-rate ground truth for a subject wearing a body-mounted camera, but we leave this as future work. Because all of the markers were not visible from a single surveying location, we mapped pairs of consecutive coordinate frames to define frame-to-frame \SE{3} transforms, and compounded them to compute the coordinates of each marker in a single navigation frame.

To compute each frame-to-frame \SE{3} transform, we surveyed the coordinates of five locations on each of the two AprilTags in the MultiStation frame: two along one edge, two along an orthogonal edge, and one at the intersection of the two edges (which defined the origin of the frame). We then used an analytic point cloud alignment procedure \cite{Umeyama1991-ws} to compute the transformation between the MultiStation frame ($\cframe{m}$) and each of the two AprilTag frames ($\cframe{i}$ and $\cframe{i+1}$). We obtained the final transform between each of the AprilTag coordinate frames by evaluating 
\vspace*{-1mm}
\begin{equation}
\Matrix{T}_{i+1, i} = \Matrix{T}_{i+1, m} \Matrix{T}^{-1}_{i, m},
\end{equation}
\vspace*{-1mm}
where $\Matrix{T} \in \text{SE(3)}$. We repeated this process for every set of adjacent AprilTags, and then compounded the transforms to compute all six marker positions in a global navigation frame.  To estimate the accuracy of the ground truth markers, we applied this procedure in the forward and backward directions to determine the loop closure error.  Our method achieved a loop closure error of 0.31 m over a path length of 130.6 m (0.24$\%$ error). For the final ground truth map, we used the position of the markers from the forward direction only, although we note that it is also possible to use pose-graph relaxation to incorporate both sets of measurements into one consistent map. 

Our accurate mapping procedure allows us to evaluate INS position error at intermediate points along the trajectory (as opposed to computing only a loop closure error based on the start and finish).  Due to symmetries within a particular trajectory, loop-closure errors can be deceptively low compared to errors at other points along the trajectory. For example, if the zero-velocity detector is set to a higher-than-optimal value, zero-velocity detection will occur slightly before and after midstance.  The INS will then underestimate the user's step length \cite{Nilsson:2012}.  However, if the user returns to the origin, the underestimation occurs in both directions along the trajectory, effectively removing its effect from the final loop closure error. By observing the position error at the furthest point from the origin, we can obtain a better quantification of error accumulation for these types of trajectories.

\subsubsection{Data Collection}
\begin{figure*}
	\centering
	\begin{subfigure}[]{0.48\textwidth}
		\includegraphics[width=\textwidth]{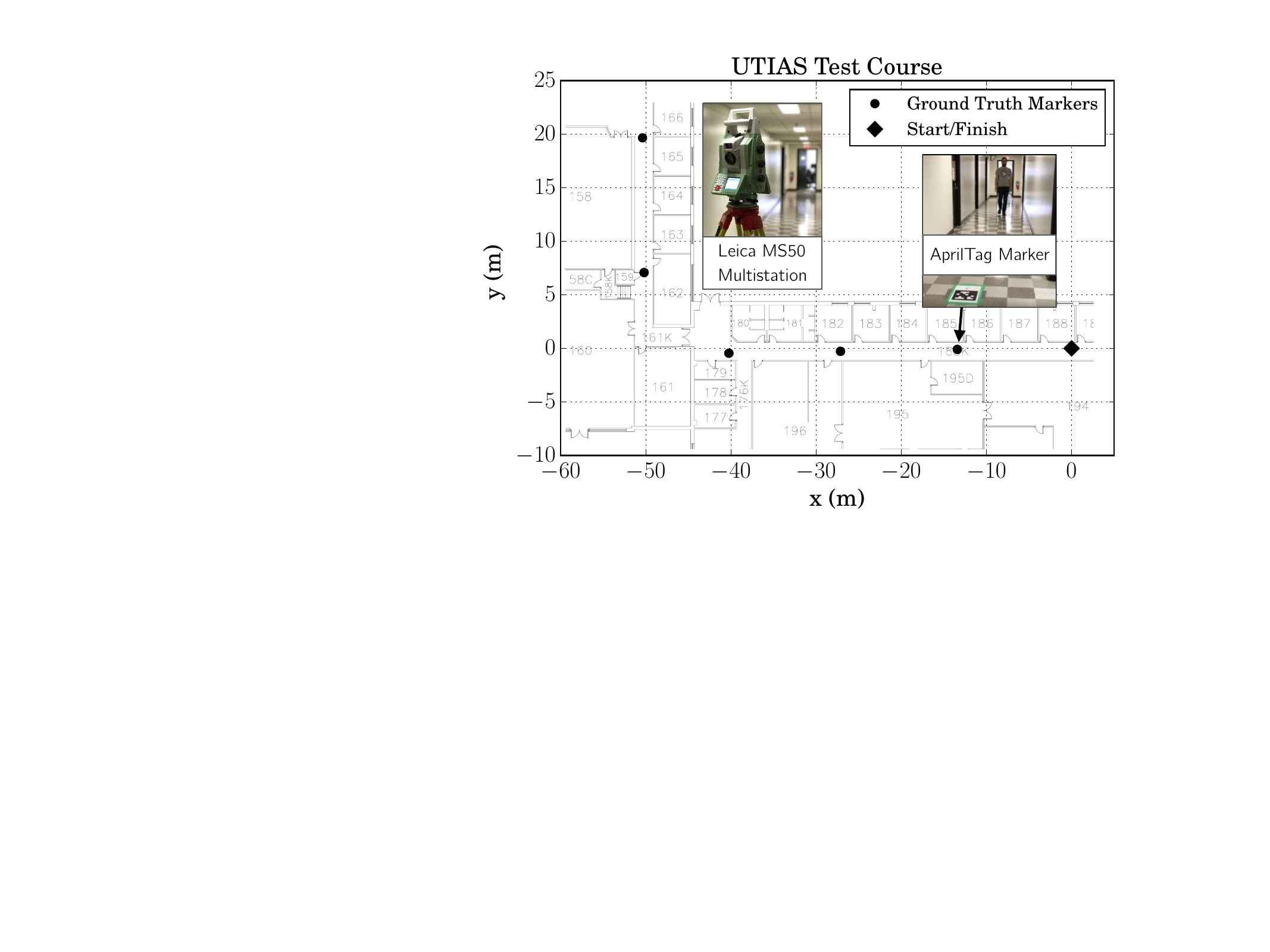}
		\caption{The UTIAS hallway course floor map with six ground truth markers (surveyed using a Leica Multistation).}
		\label{fig:hallway}
	\end{subfigure}
	~
	\begin{subfigure}[]{0.48\textwidth}
		\includegraphics[width=\textwidth]{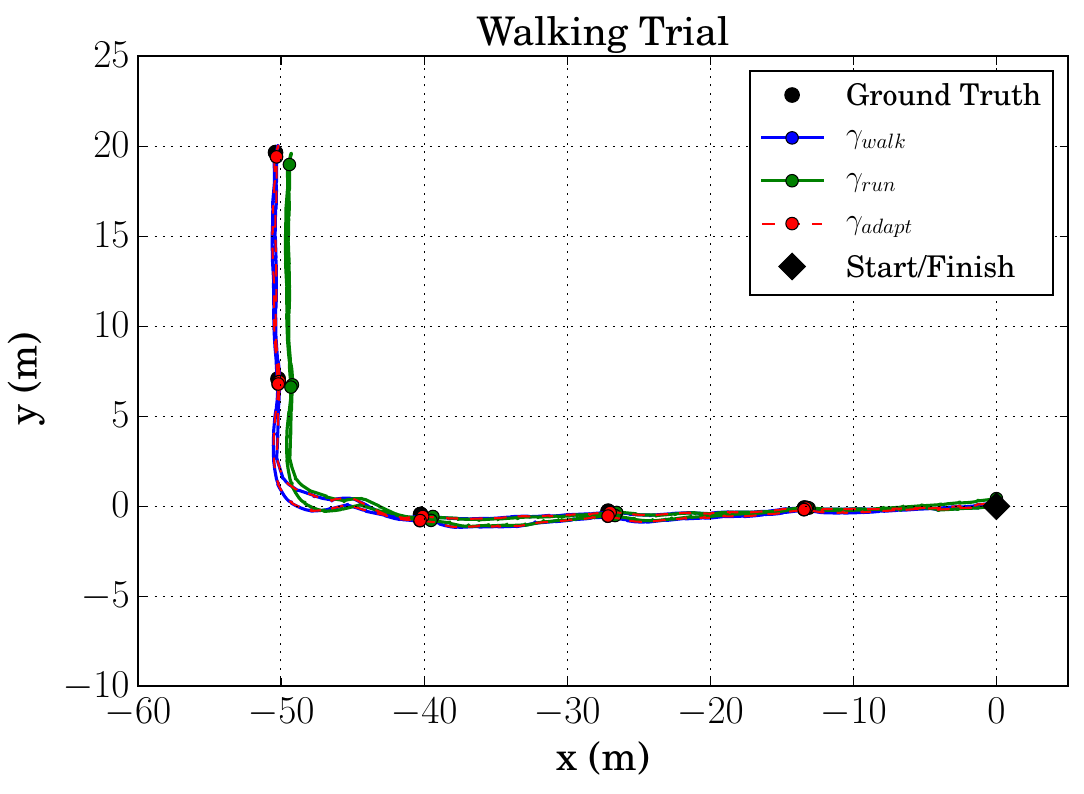}
		\caption{Pure walking motion.  Note that the adaptive threshold trajectory is visually equivalent to the walking threshold trajectory.}
		\label{hallwalk}
	\end{subfigure} \\
	\begin{subfigure}[]{0.48\textwidth}
		\includegraphics[width=\textwidth]{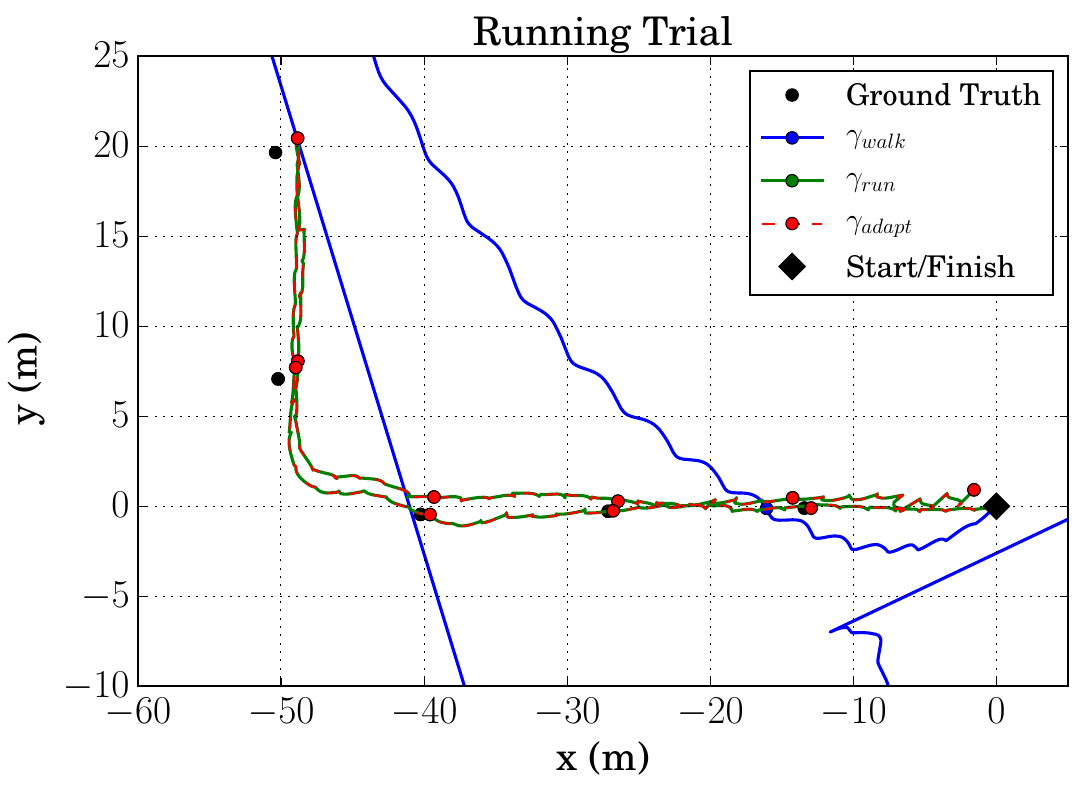}
		\caption{Pure running motion.  Note that the adaptive threshold trajectory is visually equivalent to the running threshold trajectory.}
		\label{hallrun}
	\end{subfigure}
	~
	\begin{subfigure}[]{0.48\textwidth}
		\includegraphics[width=\textwidth]{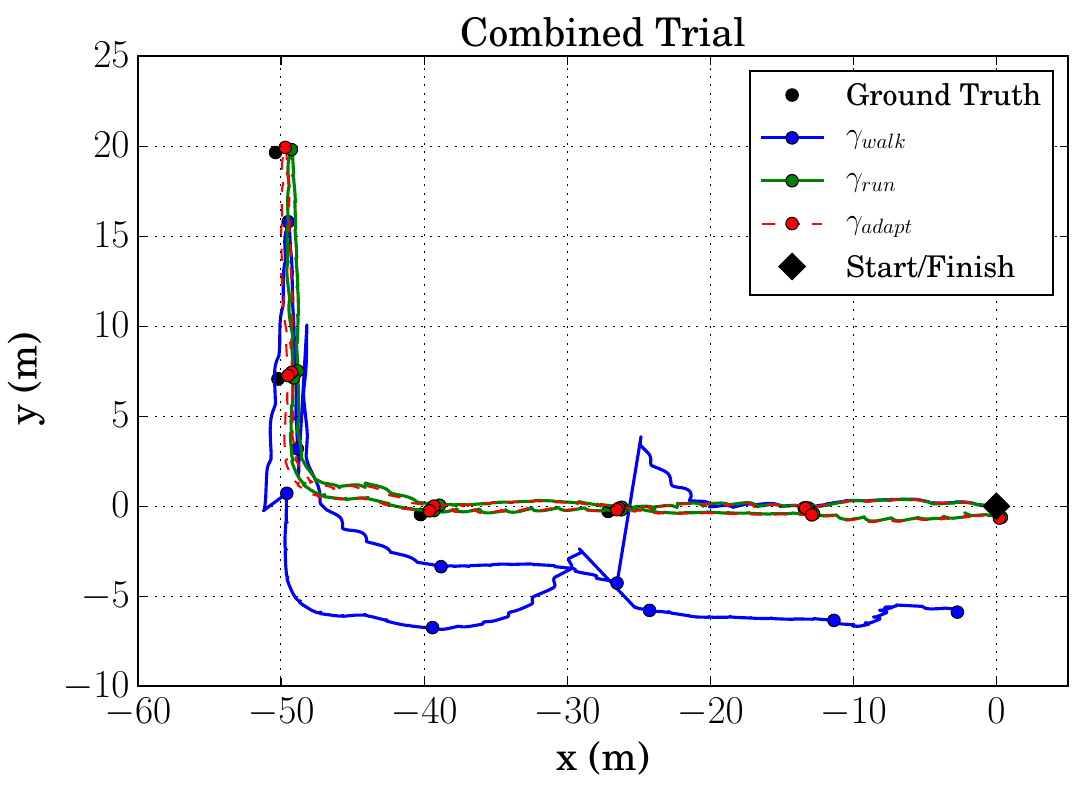}
		\caption{Combined running/walking motion. The adaptive threshold outperforms both the walking and running threshold.}
		\label{hallcomb}
	\end{subfigure}
	~
	\caption{INS position estimates using two different zero-velocity thresholds and our adaptive approach. All trajectories come from subject 4.}
	\label{fig:halltraj}
	\vspace*{-2mm}
\end{figure*}

We collected inertial data from five individuals who each recorded three walking, running, and combined running/walking motion trials along the trajectory with the surveyed ground truth.  For each trial, users started at the origin, walked through a hallway (approximately 50 m), made one $90^o$ right-handed turn, walked a further 20 m to the furthest marker, and then turned around and retraced their steps to the origin (see \Cref{fig:hallway}). At each ground truth marker, the subject pressed a handheld trigger that recorded a timestamp to facilitate temporal alignment with ground truth. For the combined trial, we instructed the users to alternate motions between every consecutive ground truth marker (beginning the trial with walking, and ending with running).

\subsubsection{Motion Classification}
We trained user-specific SVM motion classifiers with approximately one minute of walking and running data that we collected for each subject prior to evaluation.   We filtered the SVM output in order to reduce artifacts around motion transitions by applying a mean filter,
%
\[ \bar{y}_{i} =  \left\{
\begin{array}{ll}
1 & \frac{1}{W_s} \sum_{i}^{i+W_s} y_i \geq 0.2 \\
0 & \frac{1}{W_s} \sum_{i}^{i+W_s} y_i \leq 0.2, \\
\end{array} 
\right. \]
\vspace{-2mm}

\noindent where, for this work, we use $W_s = 15$. \Cref{fig:motionclass} depicts an example of the unfiltered and filtered binary motion classifier while a user alternated between walking and running.  Filtering with a threshold less than 0.5 causes the classifier to identify running rather than walking during motion transitions, reducing the likelihood that the walking threshold is applied to the first running steps when a user abruptly increases their movement speed (a case that often leads to missed midstance detection, and increased error).  We note the classification requires 1 second of data (125 IMU samples) and therefore there is a short lag behind the true motion being performed.

We compared the SVM motion classification with known ground truth (based on either the known motion class for that trial, or the handheld trigger signal for the combined trial). For the combined motion case, we note that our ground truth motion may differ from the user's true motion because users cannot instantaneously transition between walking and running.  Rather, there is a transition period, which we choose not to model in this work.
  
\subsubsection{Results}

For each trial, we determined pose estimates along the trajectory using the EKF-based INS described in \Cref{sec:overview}.  We used each of the user-specific optimized static thresholds ($\gamma_{walk}$ and $\gamma_{run}$), and also the proposed adaptive technique that applied the appropriate threshold (identified as $\gamma_{adapt}$) for the user's current motion type
%
\[ \gamma_{adapt} =  \left\{
\begin{array}{ll}
\gamma_{walk} & \bar{y}_i=0 \\
\gamma_{run} & \bar{y}_i=1. \\
\end{array} 
\right. \]
\vspace{-2mm}

To evaluate the accuracy of each thresholding method, we computed the Euclidean norm of 2D x-y position error at the furthest point of the trajectory.  Figure \ref{fig:halltraj} depicts three characteristic motion trials from subject 4, with trajectories produced by each of the three thresholding methods.

Table \ref{tab:endRMSE} summarizes the results of each thresholding method.  First, we observe that, as expected, using the optimized threshold for a given motion results in the lowest error during single-motion trials: $\gamma_{walk}$ results in a lower position error than $\gamma_{run}$ for every user during walking, and $\gamma_{run}$ results in a lower error than $\gamma_{walk}$ for every user during running.  Note that the difference in the running trials is more apparent because $\gamma_{walk}$ is too low to detect the majority of zero-velocity events while running.  Owing to the high accuracy of our classifier, using the adaptive threshold for the pure motion cases results in position errors approximately equal to that for the optimized fixed thresholds.  

In the combined case, our adaptive approach ($\gamma_{adapt}$) resulted in more accurate position estimates than $\gamma_{walk}$ or $\gamma_{run}$, with an average end-point error of 2.68 m compared to 7.35 and 3.30 m for $\gamma_{walk}$ and $\gamma_{run}$, respectively.  We note that the error reduction using $\gamma_{adapt}$ relies on the SVM being able to accurately classify motion type.  \Cref{tab:endRMSE} shows that our SVM classifier achieved accuracies greater than 95\% for pure walking and pure running trials, but a slightly lower accuracy (81\% or better) for all combined motion trials. This reduction in accuracy can be explained, in part, by the fact that our ground truth motion types switch instantaneously between markers, while our subjects often needed a few steps to make the transition. Modelling this transition is left as future work.

\begin{table}[]
		\begin{threeparttable}
\centering
\caption{SVM classification accuracy and localization error for walking, running and combined motion trials. Total roundtrip path length was approximately 130 m. We compute furthest point error at the furthest ground truth marker from the origin.}
\label{tab:endRMSE}
	\begin{tabular}{cccccc}
		\multirow{2}{*}{\textbf{Trial}} & \multirow{2}{*}{\textbf{Subject}} & \multicolumn{1}{l}{\multirow{2}{1.7cm}{\centering \textbf{SVM Accuracy (\%)}}} & \multicolumn{3}{c}{\textbf{Furthest Point Error (m)}} \\ \cline{4-6} 
		&                                   & \multicolumn{1}{l}{}                                            & $\gamma_{walk}$  & $\gamma_{run}$  & $\gamma_{adapt}$ \\ \midrule \T \T \B
		\multirow{6}{*}{Walking}        & 1                                 & 99.01                                                           & 2.17             & 3.68            & 2.24             \\
		& 2                                 & 97.93                                                           & 1.33             & 2.32            & 1.26             \\
		& 3                                 & 96.80                                                           & 0.60             & 1.49            & 0.63             \\
		& 4                                 & 98.61                                                           & 0.62             & 4.97            & 0.60             \\
		& 5                                 & 98.53                                                           & 1.84             & 3.87            & 1.81             \\
		& Mean                              & \textbf{98.18}                                                  & \textbf{1.31}    & \textbf{3.27}   & \textbf{1.31}    \\ \midrule \T \T \B
		\multirow{6}{*}{Running}        & 1                                 & 95.56                                                           & 18.37            & 3.93            & 3.54$^1$             \\
		& 2                                 & 98.41                                                           & 109.26           & 4.28            & 4.23             \\
		& 3                                 & 98.06                                                           & 98.11            & 2.90            & 2.91             \\
		& 4                                 & 95.73                                                           & 248.63           & 7.75            & 7.74             \\
		& 5                                 & 96.63                                                           & 165.57           & 2.65            & 2.67             \\
		& Mean                              & \textbf{96.88}                                                  & \textbf{127.99}  & \textbf{4.30}   & \textbf{4.22}    \\ \midrule \T \B
		\multirow{6}{*}{Run/Walk}       & 1                                 & 87.12                                                           & 7.87             & 4.77            & 4.37             \\
		& 2                                 & 81.23                                                           & 5.05             & 3.91            & 3.51             \\
		& 3                                 & 87.77                                                           & 3.14             & 1.33            & 0.96             \\
		& 4                                 & 82.39                                                           & 20.04            & 3.69            & 2.69             \\
		& 5                                 & 81.97                                                           & 0.64$^2$             & 2.79            & 1.86             \\
		& Mean                              & \textbf{84.10}                                                  & \textbf{7.35}    & \textbf{3.30}   & \textbf{2.68}    \\ \bottomrule
	\end{tabular}
		\begin{tablenotes}
			\small
			\item 1.  The mean error for the running trials is slightly lower for our adaptive approach, likely due to subject 1's slow running pace.
			\item 2. During the run/walk trial, subject 5 did not increase their running speed to the point where midstance detection using $\gamma_{walk}$ failed, resulting in a lower error than $\gamma_{adapt}$. 
		\end{tablenotes}
	\end{threeparttable}
\end{table}

\section{Conclusion \& Future Work}

We have presented an adaptive zero-velocity-aided INS that uses a motion classifier to improve tracking during sundry motion types.  We evaluated our SVM-based classifier and reported classification accuracies exceeding 90\% on a dataset of five subjects performing six different motion types. For a particular motion type, we described a method to compute optimal zero-velocity detector thresholds by maximizing an F-score given training data collected within a motion capture room. Combining the classifier with the optimal thresholds, we evaluated our final adaptive INS on a substantial indoor navigation dataset with 5.9 km of walking and running data from five different subjects. During pure walking or running motion, our system achieved localization accuracy that was equivalent to that achieved using a fixed, optimized threshold for that particular motion. In combined walking and running activities, our adaptive approach resulted in lower position errors than with either the optimized running or walking threshold alone.

These results are a proof-of-concept demonstration of the capability of an adaptive, classification-based INS.  In future work, we hope to analyze the effect of optimizing additional parameters, extend the real-time classification to other motion types and motion transitions, and to incorporate other zero-velocity detection paradigms (trained or hand crafted) that may work better for motions like crawling.

\balance
\bibliographystyle{IEEEtran}
\bibliography{zupt_ipin2017.bib}

\end{document}